\documentclass[11pt,a4paper]{article}
\usepackage{hyperref}
\pdfoutput=1
\usepackage[pdftex]{graphicx}
\usepackage{pdfpages}
\usepackage{booktabs} 
\usepackage{epstopdf}
\usepackage{graphicx}
\usepackage[position=t,singlelinecheck=off]{subfig}
\usepackage{caption}

\usepackage{color}

\usepackage[linesnumbered,ruled]{algorithm2e}
\usepackage{array,color}
\usepackage{listings}
\usepackage{graphicx,subfig}
\usepackage{amssymb,amsfonts,amsmath}
\usepackage[a4paper,left=25mm,right=20mm,top=15mm,bottom=15mm]{geometry}
\usepackage[ruled]{algorithm2e} 
\usepackage{amsmath}
\usepackage{colortbl,booktabs}

\usepackage{subfig}
\usepackage{threeparttable}
\usepackage{footnote}
\usepackage{balance}

\usepackage{diagbox}
\usepackage{epstopdf}
\usepackage{tablefootnote}
\usepackage{multirow}
\usepackage{enumitem}
\usepackage{amsfonts}
\usepackage[ruled]{algorithm2e} 

\begin{document}
\title{\bf{To Root Artificial Intelligence Deeply in Basic Science \\ for a New Generation of AI}}
\date{}
\author{\sffamily Jingan Yang$^{1,*}$, Yang Peng$^{2}$ \\
    {\sffamily\small $^1$ Institute of AI and Robotics, Hefei University of Technology Hefei 230009, China}\\
    {\sffamily\small $^2$ State Key Lab. of Software Technology $\&$ Applications, China}}
\renewcommand{\thefootnote}{\fnsymbol{footnote}}
\footnotetext[1]{Corresponding author}
\maketitle
%
\begin{abstract}
One of the ambitions of artificial intelligence is to root artificial intelligence deeply in basic science while developing brain-inspired artificial intelligence platforms that will promote new scientific discoveries. The challenges are essential to push artificial intelligence theory and applied technologies research forward. This {\color{blue}\texttt{paper}} presents the grand challenges of artificial intelligence research for the next 20 years which include:~({\textit{i}}) to explore the working mechanism of the human brain on the basis of understanding brain science, neuroscience, cognitive science, psychology and data science; ({\textit{ii}}) how is the electrical signal transmitted by the human brain? What is the coordination mechanism between brain neural electrical signals and human activities? ({\textit{iii}})~to root brain-computer interface~(BCI) and brain-muscle interface~(BMI) technologies deeply in science on {\color{blue}\texttt{human behaviour}}; ({\textit{iv}})~making research on knowledge-driven visual commonsense reasoning~(VCR), develop a new inference engine for cognitive network recognition~(CNR); ({\textit{v}})~to develop high-precision, multi-modal intelligent perceptrons; ({\textit{vi}})~investigating intelligent reasoning and fast decision-making systems based on knowledge graph~(KG). We believe that the frontier theory innovation of AI, knowledge-driven modeling methodologies for commonsense reasoning, revolutionary innovation and breakthroughs of the novel algorithms and new technologies in AI, and developing {\color{blue}\texttt{responsible}} AI should be the main research strategies of AI scientists in the future.

\end{abstract}
\maketitle   
\setcounter{page}{1}
\section{Introduction}
\subsection{Essence of artificial intelligence}
{{The intention of artificial intelligence is to use computers to simulate human intelligence, so that machines have consciousness, can think, learn autonomously, can listen, speak, understand language, and solve problems based on intelligent reasoning and rapid decision-making. It specifically related to: perception, computation, cognition, and innovative intelligence. Perception includes vision, hearing, touch, and taste, and the whole body of a person can be seen as a perceptron. Computing capability is mainly reflected in memory and computing power. Computers are far superior to humans in this respect. For cognitive intelligence, it is mainly reflected in understanding, thinking, autonomous learning, reasoning and decision-making. For example, cognitive robots, natural language understanding, knowledge-based reasoning and decision-making. Creative intelligence is mainly reflected in the autonomous learning in the unknown environments using commonsense and experience through imagination for building models and algorithms, designing and experimentally verifying the intelligent process. The main goal of artificial intelligence research focuses on machines that can achieve complex tasks that usually require human intelligence to complete. Artificial intelligence is becoming a basic, leading technology that promotes the development and transformation of various industries and fields{\bf\color{purple}{\cite{wu02}}}. An artificial intelligence machine can be understood as a type of machine that has the ability to autonomously and interactively execute and complete anthropomorphic tasks in various environments. The development of AI is creating new opportunities to solve challenging, real-world problems. AI, big data, 5G, Internet of Things~(IoT), and other technologies will be an engine of a new round of industrial revolution.}}
\subsection{Recent progress in artificial intelligence}
Artificial intelligence is not a novel field, but a branch of computer science that involves the multidisciplinary intersection of basic theory, engineering design, and applications. Artificial intelligence focuses on building intelligent machines with functions that mimic the intelligence of the human brain. These machines possess thinking activities similar to human perception, recognition, understanding, reasoning, judgment, proof, autonomous learning, and problem solving. They can work like humans and respond to the various things and environments.
The discipline of artificial intelligence was born in the Dartmouth Conference in 1956. At this conference, John McCarthy first used the concept of {{artificial intelligence}}. In 1957, Rosenblatt proposed the neural network perceptron, which is the basis of {{Support Vector Machine}}~(SVM) and Neural Network so that AI research reached the first climax. In 1986, David proposed that BP algorithm made large-scale neural network training possible, and AI ushered in the second golden period. At the end of the 1990s, with the in-depth study of machine learning, evolutionary computing, evolutionary robots, and swarm intelligence, Later, the invention of powerful, simple, and versatile deep learning{\bf\color{purple}{\cite{lecun01}}}, the successful development of self-driving cars, AI research culminates. From 2012, Google's image recognition error rate was significantly improved, and also the machine's image recognition accuracy was significantly improved and approached human level. In 2016, Alphago defeated the world Go champion. In 2017, AlphaZero defeated Alphago, and DeepMind tried StarCraft games. The rapid development of AI has pushed artificial intelligence to a new climax. Since then, human beings have entered an artificial intelligence era. However, whether artificial intelligence will defeat humans and dominate the world? The answer is {\bf{"no"}}. Artificial intelligence will always serve humanity.
\subsection{Challenges and countermeasures}
Artificial intelligence has approached or reached human level in some applications in the perception intelligence fields{\bf\color{purple}{\cite{zhuang01}}}, but research in the cognitive intelligence fields {\texttt{is still in its infancy}}. We aim to make progress in cognitive intelligence, that is, artificial intelligence inspired by the brain, which understands {\emph{how the brain makes the mind, how the brain works and communicates, how brain electrical signals control human activities, and how to build intelligent machines}}. Therefore, we should acquire knowledge and experience from brain science, neuroscience, cognitive psychology, and human commonsense, and combine dynamic knowledge graph, causal reasoning, autonomous learning and implicit knowledge to build an effective mechanism for stable knowledge acquisition and knowledge representation, so that the machines can accurately understand and use knowledge for achieving a key breakthrough from perception intelligence to cognitive intelligence. This {\texttt{paper}} proposes six challenges for artificial intelligence research: ({\textit{i}}) brain cognitive functions and brain-inspired intelligent computing models: through the research of brain science, neuroscience, cognitive science, psychology and data science, understand the working mechanism of the human brain and build brain-inspired machine learning models; ({\textit{ii}}) How are the human brain electrical signals transmitted? What is the mechanism of the coordination of brain electrical signals and human activity? ({\textit{iii}}) In-depth study of brain-computer interface theory and key technologies; ({\textit{iv}})~Research on knowledge-based visual commonsense reasoning, building a knowledge-driven commonsense library, developing a new inference engine "cognitive network recognition~(CNR)"; ({\textit{v}}) research and development of high-precision, multimodality intelligent perceptrons; ({\textit{vi}})~Key technology breakthroughs in intelligent reasoning and rapid decision-making based on knowledge graph (KG). We believe that frontier basic theory and innovation of artificial intelligence, knowledge-driven commonsense reasoning, a novel algorithm and innovation, technology application and innovation, AI talent training and development, AI ethics and morality should be the main research strategies of AI scientists in the future.
\section{Current State-of-the-Art of Artificial Intelligence}
The research progress of artificial intelligence is tortuous, that is, there are climaxes and troughs. {{\color{blue}{The first trough}} of AI research: 1974-1980.
{{\color{blue}The second trough}} of AI appeared in around 1987-1993. This trough comes from the craze for the expert systems around 1980. When the enthusiasm for expert system research is rising, and the funding reduction for AI research led to this trough, and soon people are disappointed. It is estimated that {{\color{blue}The next trough}} of artificial intelligence research may appear in the next 5-10 years.

In recent years, the development momentum of AI is unstoppable. In fact, machine learning has become an indispensable part of up-to-date AI, so that the terms "artificial intelligence" and "machine learning" are sometimes used interchangeably. However, this may lead to ambiguity in professional fields and inaccurate use of language. The best way to understand it is that machine learning represents {\emph{the latest technology}} in AI fields, and deep learning reflects {\emph{the rapid development of machine learning technology}}{\bf\color{purple}{\cite{lin01}}}, but deep learning does not represent artificial intelligence. Moreover, the unexplainability~(black box) and poor stability of deep learning are current international research hotspots. Both directly affect the universality and robustness of artificial intelligence. In recent years, computer vision, autonomous robots, self-driving cars, and natural language processing~(NLP) have become exciting sources of innovation. This benefits from the development of machine learning~(ML) and deep learning~(DL). However, data-driven depth learning algorithms are not inspired from brain science. Although convolutional neural network~(CNN) directly simulates the mechanisms of the visual cortex of the brain, uses hierarchical coding, local network connection, and pooling, which are directly related to the biological mechanisms, but uses a completely non-biological training methods, so that some people miscalled {\emph{the training of the CNN}} "deep learning".

At present, artificial intelligence research and development has been selected as a national strategy for science and technology innovation by many countries. Artificial intelligence colleges under construction, artificial intelligence majors, and various artificial intelligence institutions are springing up. The development of the artificial intelligence industry is also very prosperous. But less attention has been paid to the frontier basic theories of artificial intelligence. If you want to make a breakthroughs in the frontier basic theories and key technologies of the AI, it may depend on the breakthroughs made in brain science, neuroscience, psychology, cognitive science, and materials science~(the robot entities that include software robots~(Softrobots) sensing and driving{\bf\color{purple}{\cite{yang01,laschi01}}}, muscle, nerve, retina~(Macula, etc.) and medical and other disciplines. Therefore, we must make research on the frontier basic theories and key technologies of the AI in a down-to-earth manner, explore the key technologies in artificial intelligence and its applications, and find ways to crack the bottleneck of key technologies.
\section{Different Stages of Artificial Intelligence Research}
Artificial intelligence is usually divided into three categories: {\bf{The first category}} is also the most common weak AI, that is, artificial intelligence, also known as {\emph{narrow intelligence}}. Narrow intelligence focuses on completing {\emph{specific tasks}}, and represents almost all existing AI systems, of course, also include {\emph{knowledge engineering and expert systems}}. To complete the tasks at this stage is mainly based on computer science, modern physics, basic mathematics, and computational neuroscience. Based on mechanisms discovered in various physical experiments, we use mathematics to formulate the problems and construct computational models which generate various computer-implemented algorithms based on the computational models, find the optimal algorithms and the optimal solution of the algorithms. At present, statistics is widely used in such artificial intelligence like big data processing, computer vision, image feature recognition and understanding, scene and target tracking in video, recognition and understanding, speech recognition, medical diagnosis, and machine learning, etc. The key problem of current artificial intelligence is the lack of {\emph{human knowledge and experience because}} it is difficult to accurately express {\emph{human knowledge and experience}}.

{\bf{The second type of AI}} is also known as an artificial general intelligence(AGI){\bf\color{purple}\cite{legg01}}. It means that the systems have the ability to complete the tasks like humans (with human intelligence), that is, to complete human-level tasks. In fact, the goal of artificial intelligence research is to build a universal artificial intelligence systems, and brain-like computing is an important way to achieve it. However, such systems have not yet been realized. In 2013, scientists in computer and neuroscience and a scientific team consisting of experts, medical experts and researchers released predictions that the probability of developing a real~(general) artificial intelligence~(AGI) between 2040 and 2050 is about {\color{red}50\%}. The realization of general artificial intelligence AGI may mainly depend on research breakthroughs in the disciplines of brain science, computational neuroscience, psychology, cognitive science, materials science, and medicine. The experiments reveal the nature of human thinking, perception~(vision, hearing, touch, and taste), memory and storage, and neurons transmitting and processing information mechanisms. Based on the discovered mechanisms, various types of calculation models are constructed and various algorithms are generated, such as thinking models, perceptual models, memory models, storage models, and computer-implemented algorithms, etc. Whether the general artificial intelligence AGI can be truly realized, scientists still need to make great efforts to study and expose the mysteries of human brain intelligence, and build {\emph{brain-like AGI systems}}. In essence, the closer AGI research is to the ultimate algorithm of human intelligence, the better artificial intelligence algorithms can be found, and the stronger the AGI system will be. Looking forward to the realization of the AGI systems as soon as possible.

In order to realize {\emph{true artificial intelligence}} and {\emph{the unique human-like intelligence}} based on computers, so that the machines would have the functions of seeing, listening, speaking, thinking, understanding language, autonomous learning, reasoning and rapid decision-making, and problem solving and the possibility of creating computer simulation of human brain learning ability. So we must understand the brain's storage, memory, mechanisms of functions such as perception, learning, reasoning, etc., to create a complex mathematical model and effective algorithms implemented with computers, anthropomorphic brain 8.6 billion neurons and about 100 trillion synapses, and the way they handle and the mechanism of transmission of information, it is still a long, there are many difficulties to overcome.

{\bf{The third type of artificial intelligence}}, also called {\emph{strong}} artificial intelligence, should be {\emph{universal and interpretable}}. These AI systems can be understood as not {\emph{simulating}} human intelligence, but {\emph{surpassing}} human intelligence, and even {\emph{smarter than humans}}. This may be the ultimate goal of AI research.
\newline{}
\section{To Root Artificial Intelligence Deeply in Basic Sciences}
The goal of artificial intelligence is to build intelligent machines with human-level control. These machines have functions similar to human perception, recognition, understanding, reasoning, judgment, proof, autonomous learning, and problem solving, and can work like humans. But what the current AI lacks is a good learning model similar to human learning, and the key to machine learning is to learn with human learning mechanisms. In order to achieve this goal, we root artificial intelligence deeply in the basic sciences for developing a new generation of AI.~This paper proposes the frontier theoretical research in AI fields and hope to have {\emph{revolutionary breakthroughs}} in the following areas.
\subsection{Human-inspired machine learning mechanisms}
Humans and animals can learn to see, listen, perceive, act, and communicate with an efficiency that machine learning method can not approach.

What the current AI lacks is a good model from human learning. The key to machine learning must be to imitate or duplicate human learning mechanisms. To achieve human-like machine learning, it must be done through brain science and neuroscience, cognitive science, psychology, and data science, understand the working mechanism of the brain, and build brain-inspired machine learning models.

At present, computers have considerable storage capacity, but their ability to recognize patterns, learning to learn, and understand information is far from being able to match human brain. To solve this problem, we must reveal the true connection mechanism of brain neurons and better understand how brain neurons are connected. Once we can understand the real connection of brain neurons and its mechanisms of quickly transmitting and processing information, autonomous learning, fast decision-making, and its commands to sensitive actions, we can develop more sophisticated and advanced artificial intelligence systems.

The artificial intelligence foundation of intelligent machines should make research on {\texttt{learning how to learn}}, "combining advanced pattern recognition with model-based reasoning" and building a knowledge-driven commonsense database.

The AI systems which know their limitations and know how to seek help may exceed current training and knowledge acquisition methods. These systems know how to interact, how to seek help, how to recover from failures, and how to become smarter. AI and robots systems that model their own components and operations are essential for adaptation and development. We need AI that can detect own sub-components, model their operations and modify these models when the structure changes. The four-legged robots developed by Bongard {\textit{et al}}~{\bf\color{purple}{\cite{bongard01}}} can find and repair their damaged components by themselves, and autonomously learn to use these components dynamically in motion.

Deep learning is one of the fastest-growing machine learning methods. Deep learning methods has achieved a significant improvement of object recognition accuracy using hierarchical pattern recognition. This model maintains the consistency of information at every level of the hierarchy. The new machine learning algorithms are unprecedented for data combined with access to cheap and powerful computing hardwares. The results of solving AI problems in the narrow category have made progress, and many people think that we are on the brink of solving intelligence--intelligent in all aspects, and people understand it still poor. However, we have a long way to go to replicate and surpass all aspects of intelligence inspired by human beings. Combining advanced pattern recognition with model-based reasoning to build beyond statistical correlation and start to potential interdisciplinary mechanism and system dynamics for inference. Meta-cognitive Learning (MCL) or learning how to learn new things is not only important for large training data sets, but also for restricted data, it is essential new AI capabilities. The challenge is how to learn dynamically to adapt to a dynamic and unknown environment. Based on understanding neuroscience on the human and hippocampus, we has developed a promising method in this field as a predictive graph of new situations. AI is able to learn complex tasks by itself and with minimal initial training data is essential for next-generation systems. Most of machine learning systems are data intensive, require a lot of data to learn complex tasks. Performance comparison curves of data-driven machine learning algorithms are shown in~(Fig.~{\bf\color{purple}\ref{fig01}}), and DeepMind's Alpha-Go Zero system is capable of self-learning and playing Go far beyond Go's world champion. This is an impressive example. We think that understanding deep learning mechanism will not only enable us to make more intelligent machines, but will also help us understand {\emph{human intelligence and the mechanisms of human learning}}.
\begin{figure*}[!ht]
\centering
\includegraphics[width=12.8cm,height=7.2cm]{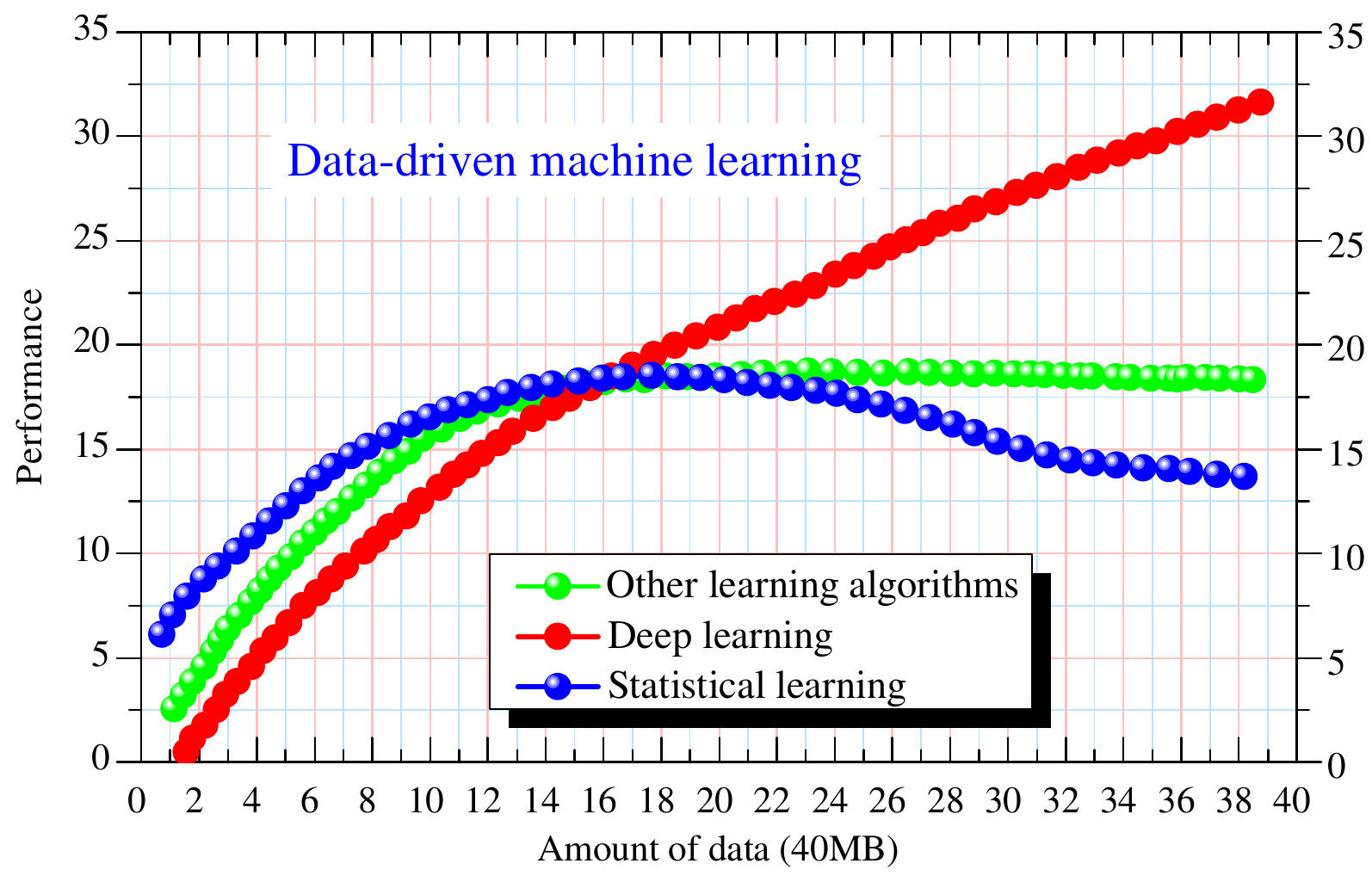}
\caption{{\bf{Performance comparison curves of data-driven machine learning algorithms.}} {\bf{a}}) statistical method for machine learning~({\color{blue}blue}); {\bf{b}}) deep learning method~({\color{red}red}); {\bf{c}}) other learning algorithms, such as reinforcement
learning{\bf\color{purple}\cite{yang02}}, etc.~({\color{green}green}))}
\label{fig01}
\end{figure*}
\subsection{Building a brain-inspired robot model}
So far, we have no {\emph{a real brain-inspired robot model with human-level control}}. Therefore, we shall deeply root robotics in science and construct a brain-inspired robot~(cognitive robot) model with cognitive functions based on the collaboration of the central nervous system and the peripheral nervous system that imitates human behavior and activities and has {\emph{autonomous learning, fast decision making, and sensitive action abilities}}. At the same time, we will also develop high-precision multi-modal cognitive robotics with {\emph{vision, audition, national language processing}}~(NLP), {\emph{thinking capabilities}}, {\emph{build brain-controlled  and mind-controlled robotics systems and its interface, platform, and experimental verification systems}}.
\subsection{Developing bioinspired hybrid robots}
{{The graceful and agile movement of animals is the result of complex interactions among many components including {\emph{central and peripheral nervous systems, musculoskeletal systems, and the environment, etc.}} The goal of biorobotics is from biological principles inspired by{\bf\color{purple}\cite{yang01}}, design robots that can match the agility of people and animals, and use them as scientific tools to study the adaptive behavior of animals. Therefore, transform basic biological principles into engineering design rules, build integrated bio-hybrid robots and bio-inspired humanoid robots~(Fig.~{\bf\color{blue}\ref{efig:01}}), break down the robot's active components into synthetic structures to create robots with performance similar to {\texttt{natural system}}{\bf\color{purple}\cite{ijspeert01,chien01}}.}}
\subsection{Knowledge-driven commonsense reasoning}
The biggest problem of artificial intelligence machines is that they lack {\texttt{common sense}}. Machines do not have common sense, and many problems will be difficult to understand, and for humans, understanding these issues is very simple. This is the importance of common sense. It is difficult to construct a common sense database based on data, because common sense is often not represented by raw data, and common sense AI usually requires a lot of well-thought-out annotated training data. Perhaps the most promising in the future is to establish a common sense database (CSD) based on {\texttt{knowledge and logical reasoning}}. In the early stage of artificial intelligence research, scientists paid attention to commonsense knowledge. With the progress of artificial intelligence research, people pay more attention to algorithms and ignore common sense knowledge, so that advanced artificial intelligence systems still lack the common sense that most 10-year-old children have. We hope to change this situation as soon as possible to achieve a major breakthrough in this field. Until this problem is solved, artificial intelligence is limited to narrow applications, not {\texttt{general}} artificial intelligence~(AGI), not even {\texttt{strong}} artificial intelligence~(SAI). Like most machine learning methods, to develop a universal artificial intelligence, we should first understand the function of the human brain better, because if you do not know how the brain works, you can't simulate it, let alone copy it.

Commonsense reasoning and commonsense knowledge are very important in artificial intelligence{\bf\color{purple}\cite{zhou02}}, Davis and Marcus (Marcus) proposed that intelligent machines do not need to directly copy human cognition, but need a better understanding of human common sense to complete various tasks that require human intelligence to perform~{\bf\color{purple}\cite{davis01}}. Paul Allen's Artificial Intelligence Institute~(AI2) proposes complex methods to solve common sense problems: integrates machine translation and reasoning, natural language understanding, computer vision, and crowdsourcing technology , Created a new and extensive basic knowledge source{\bf\color{purple}\cite{tandon02}} for AI systems. At the same time, Allen revealed a big problem of the current state of deep learning technology through analogy. Despite the strong intelligence of AI based on deep learning, when "AlphaGo defeated the world number one Go champion in 2016, the program did not know that Go was a board game". Therefore, only knowledge-driven machine learning is the shorter path to achieve commonsense reasoning(Fig.~{\bf\color{purple}\ref{fig02}}). The AI ​​system can recognize objects in less than a second, mimic human voices and recommend new music, but most machines are "intelligent" and lack the most basic understanding of everyday objects and movements, in other words, lack of common sense. DARPA and the Allen Institute of Artificial Intelligence collaborate to seek high-level cognitive and common-sense understanding of the world depicted by images{\bf\color{purple}\cite{zellers02}}. Visual commonsense reasoning programs from recognition to cognition aims to define problems and facilitate problem solving. Although no one expects to "solve" the problem within a few years. However, if AI wants to have brain-like functions, then a brain needs to be cultivated, and the role of the brain is much more than just performing classification tasks quickly

We emphasize here that commonsense knowledge~(CSK) has made {\color{blue}people and machines smarter}. But the process of implanting commonsense reasoning (CSR) into computers is obviously a long one. To achieve this goal, a knowledge-driven approach provides shortcuts while facing the need for more challenging and creative strategies.
\begin{figure*}[!ht]
\centering
\includegraphics[width=12.8cm,height=7.2cm]{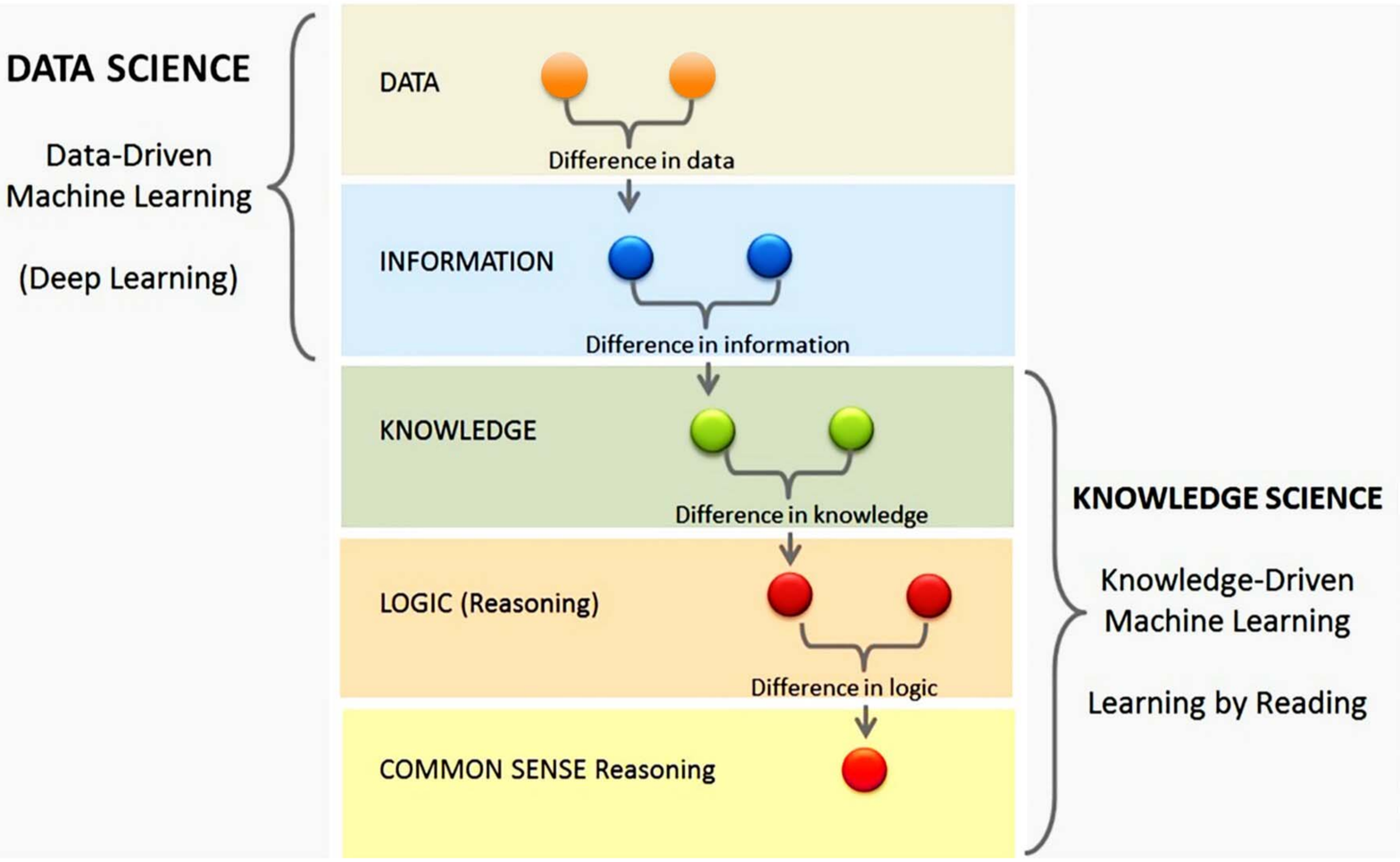}
\caption{{\bf{From data to common sense reasoning.}} Detecting differences between data will create information. The same hierarchy applies to knowledge, logical and common sense reasoning. If the differences are not detected, there will be no information, knowledge, logic or common sense reasoning. This is the most basic premise for processing intelligence. For a computer to operate at the "common sense" level, the following issues must be solved: ({\color{red}1}) obtaining information from available data, ({\color{red}2}) obtain knowledge from available information, ({\color{red}3}) derive logic from available knowledge, and ({\color{red}4}) generate common sense(reasoning) from logic (reasoning). The question is how to shorten this path for finding a feasible common sense reasoning best solution?}
\label{fig02}
\end{figure*}
\subsection{How do brain nerve electrical signals control human activities?}
{{In order to build an artificial intelligence system that can {\emph{transmit, recognize, process, and understand information as fast as the human brain}}, we must study:~({\emph{{i}}}) how is transmitted the nerve electrical signals in the brain? ({\emph{ii}}) What is the coordination mechanism between brain nerve electrical signals and human activities?
According to neuroscientists, the {\emph{"storage capacity"}} of the human brain is between 10 and 100 megabytes, and it is estimated that this number is close to 2.5PB. The human brain has a huge range of work including data analysis, pattern recognition, autonomous learning, etc. The ability to store and retain information is only certain processes that the brain goes through every day. {\bf{For example}}, people only need to look at a car a few times, they can recognize this type of cars. However, artificial intelligence systems may need to process hundreds of car samples before they can be retained enough identification information. Why is the human brain so efficient? On the contrary, although the brain functions are quite complex and information in the brain is processed massively in parallel, the human brain's ability to process the same type of big data cannot match a computer;~({\emph{{iii}}}) brain science, neuroscience, engineering science, applied sciences as well as molecular and cell biology research jointly records activities in the visual cortex of the brain, which will help understand how neurons are connected to each other. The ultimate goal is to create a more accurate and complex artificial intelligence system, that is, through these studies, we can create a computer system that can quickly and accurately understand, interpret fast, autonomously analyze and learn like humans;~({\emph{{iv}}}) These systems can be used to detect network intrusions, machine learning, self-driving cars, read MRI or complete any tasks reserved for the human brain. {\emph{The study will generate more than 1PB (1PB=1024TB) data, break through the boundaries of brain science research, expand the possible research scope of computer science}}, and make great progress in creating new data management, high-performance computing, autonomous robotics, computer vision, and network analysis, etc.}}

\subsection{Brain-computer interface theory and technology}
Brain-computer interface consists of three parts: {\texttt{brain + machine + interface}}. Where {\texttt{brain}} means the organic life brain or nervous systemas, {\texttt{machine}} can be regarded as any processing or computing device, and {\texttt{interface}} is considered as intermediary components for information exchanging. The brain-computer interface is usually divided into {\texttt{invasive, non-invasive}}, and {\texttt{semi-invasive}} according to {\texttt{intrusive}}. Also, BCI can be seen as a device that can translate neuron information into computer instructions{\bf\color{purple}\cite{donoghue01,zander01}}, these instructions can control external software or hardware, such as a computer or robotic arm~(Fig.~{\bf\color{purple}\ref{fig03}}). Techniques for converting human brain functions to cognitive robots are known as brain-to-machine or brain-to-robot interface technologies, that is, brain-inspired cognitive robots. Therefore, we must develop advanced brain function theory, build computing models for cognitive robots inspired by human cognitive mechanisms, and develop new design principles and technologies for brain-machine interfaces. BCIs can be used to establish direct online communication between brain and machines, and directly use brain activity to control computers or external devices without peripherals. BCIs enable patients with paralysis to communicate and control robotic prostheses via adjustment of the somatic motor nervous system and restore neural function during rehabilitation. It is an important application~(Fig.~{\bf\color{purple}\ref{fig04}}). BCIs can be divided into {\texttt{active, reactive or passive}}{\bf\color{purple}\cite{chaudhary01}}. The output of active BCIs come from the brain activity, and brain activities are directly and consciously controlled by the users, without having to rely on external events to control the applications. In reactive BCI, the output comes from brain activities that respond to specific external stimuli. Passive BCI is a relatively new concept, and its output comes from any brain activity that does not have a voluntary control purpose to enrich human-computer interaction and related actual users implicit information about the state. Both invasive and non-invasive methods are used to record brain activity. Intrusive methods capture neural activity in the brain through intracortical neural interfaces with microelectrode arrays, which capture {{\color{blue}Spike}} signals and Local Fields Potential~({{\color{blue}LFP}}), or cortical surface ElectrocortiCoGraphy~({{\color{blue}ECoG}}), can provide higher temporal and spatial resolution, and has good immunity to artifacts. Non-invasive BCIs do not require surgical implantation. Typical signals include slow cortex potential, sensory motor rhythm, an event-related potential brain-computer interface ({\color{blue}P300}), steady state visual evoked potential~({\color{blue}SSVEP}), (event-ralated desynchronization)~{\color{blue}ERD/ERS}~(event-related synchronization). An error-related potential(ErrP) and an event-related activity in the human electroencephalography~({\color{blue}EEG}) are used as an intrinsically generated implicit feedback~(reward) for reinforcement learning. The EEG-based human feedback in RL can successfully used to implicitly improve gesture-based robot control during human-robot interaction{\bf\color{purple}\cite{kim01,yang02}}. The signal strengths of EEG, ECoG, LFP and spikes extracted from different brain-computer interfaces vary greatly (Fig.~{\bf\color{purple}\ref{fig05}}). We commonly use assessment methods include {\color{blue}fMRI}~(functional magnetic resonance imaging), {\color{blue}fNIRS} (functional near-infrared spectroscopy), {\color{blue}MEG}~(Magnetoencephalography), and {{{\color{blue}EEG}}~(electroencephalography)}{\bf\color{purple}\cite{aljalal01}}.
\begin{figure*}[!ht]
\centering
\includegraphics[width=12.8cm,height=7.8cm]{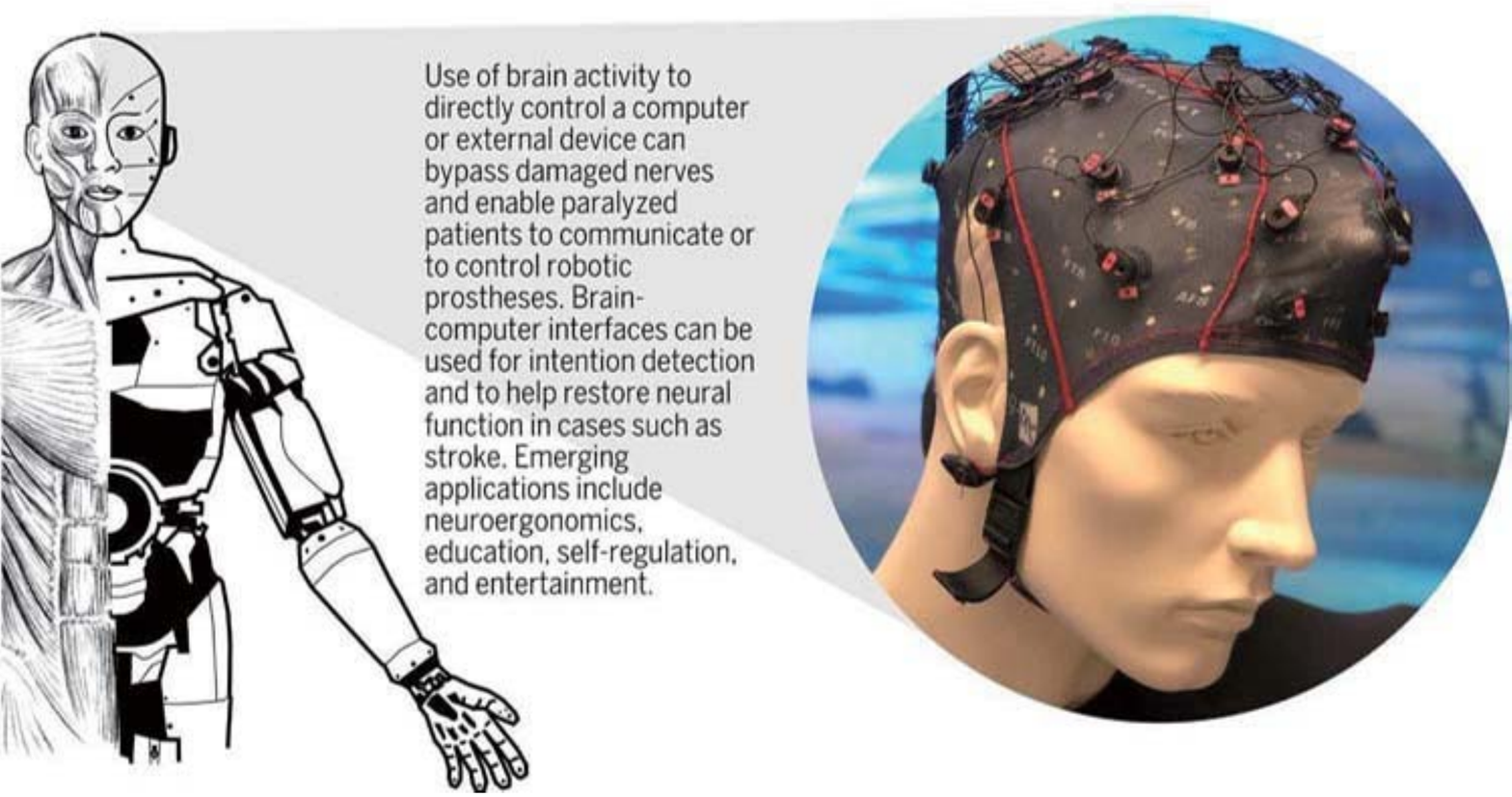}
\caption{{\bf{Use of brain activity to control external devices.}} Human brain activity is directly used to control external devices such as computers or robotic arms (cited from Science Robotics.aar7650, 2018)}
\label{fig03}
\end{figure*}
\begin{figure*}[!ht]
\centering
\includegraphics[width=12.8cm,height=7.8cm]{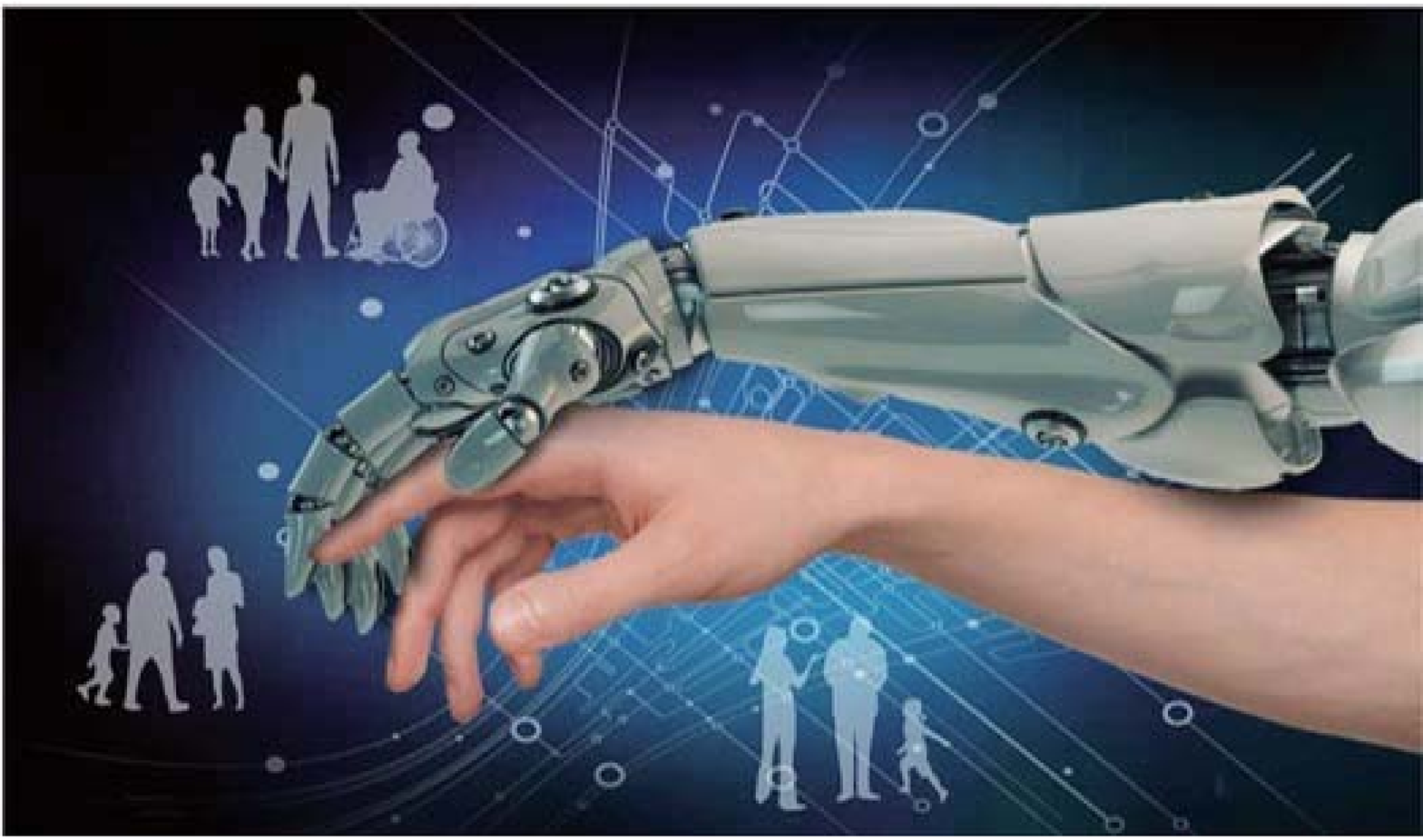}
\caption{{\bf{Brain-Computer Interface.}} BCI enables paralyzed patients to communicate with and control robotic prostheses prosthetics and is widely used in neurological rehabilitation (cited from Science Robotics.aar7650, 2018)}
\label{fig04}
\end{figure*}
\begin{figure*}[!ht]
\centering
\includegraphics[width=12.8cm,height=6.8cm]{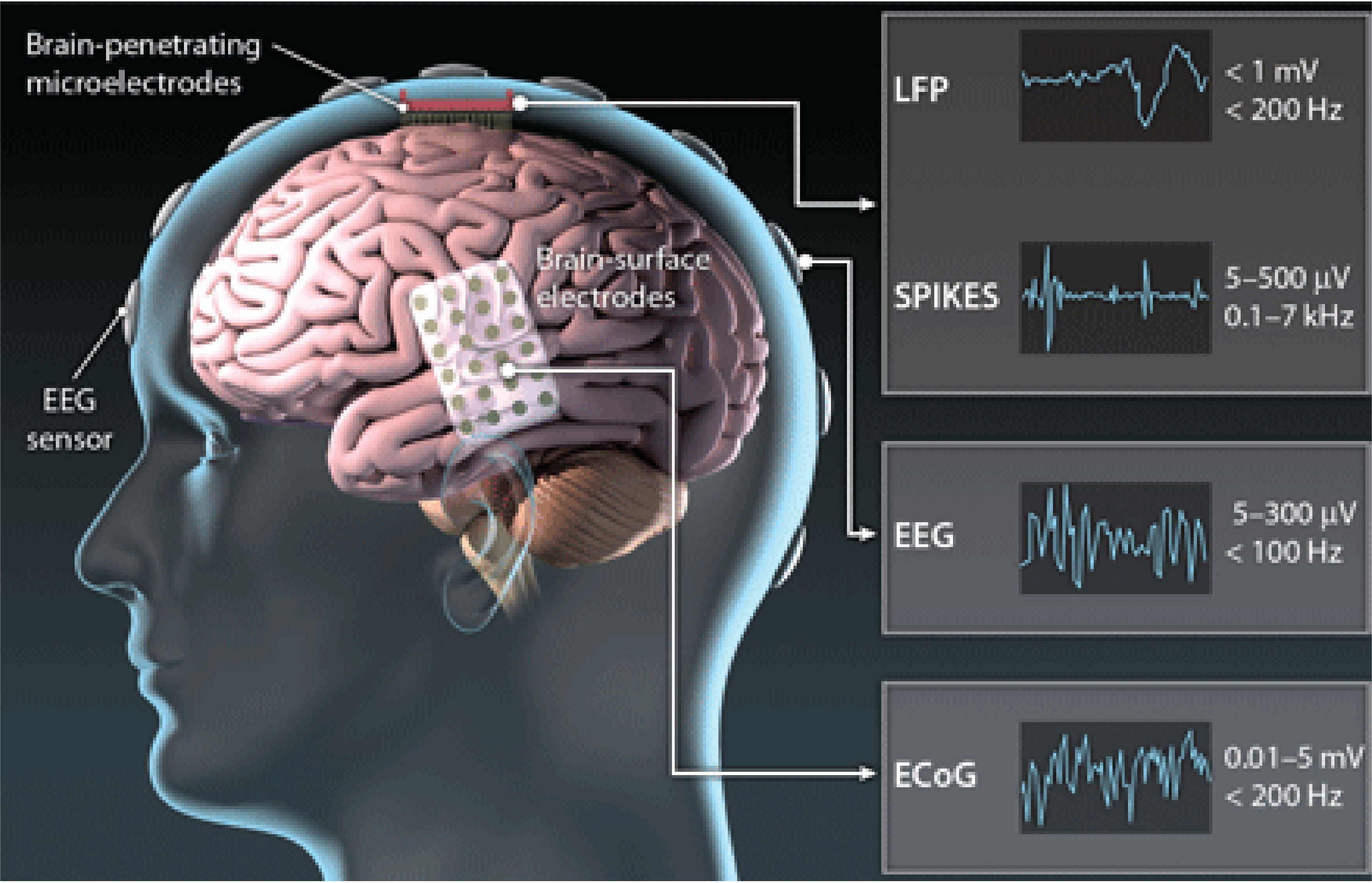}
\caption{{\bf{Extraction of EEG, ECoG, LFP and spikes.}} The signal strength obtained from different brain-computer interfaces varies greatly (cited from Neuralink, 2018)}
\label{fig05}
\end{figure*}
\subsection{Developing high-precision, multi-modal intelligent sensors}
The human body's perception of the external world is mainly through the eyes (\texttt{vision}), ears (\texttt{hearing}), tongue (\texttt{taste}), and skin (\texttt{tactile}). It can be said that whole human body can be seen as an ingenious sensor for obtaining various information. From my decades of research on vision including human vision theory, retinal vision, biological vision, computer vision, machine vision, qualitative vision, and quantitative vision, it can be concluded that the visual information obtained by normal human eyes is qualitative and complete, and visual information processing is massively parallel and extremely fast. Machine vision captures pixel-level visual information, data-driven, quantitative, incomplete{\bf\color{purple}\cite{herath02}}.

\noindent {\bf{Example 1}}: when a child encounters a stranger by accident, he has preliminary impressions of height, weight, appearance of this stranger. While they met again and noticed the person's face, walking posture, gait, and preliminary information. The third encounter may focus on the stranger's facial skin color, facial features, hairstyle, and other characteristic details. After the three encounters, the child's brain remembered the complete information of the stranger. When he met this stranger for the fourth time, the stranger can be identified or proved. However, it is not so easy for the machine, it may need to learn dozens of times, including positive, side, back, walking posture and gait, etc, even to learn hundreds of times to accurately identify. {\texttt{Compared with machines, one of the most obvious differences is the ability of people and animals to learn}}\\ {\texttt{from very few examples}}.

\noindent {\bf{Example 2}}: A teenager wants to cross the road in an emergency, He find that The green traffic light is still on for 9 seconds. As usual, It takes 15 seconds to pass safely. He speeds up walking and his eyes quickly scan both ends of the road from left$\rightarrow$right, then from right$\rightarrow$left of the road, and finally pass safely. So how to transmit such a large amount of visual information to big brain? How do brain neurons convert this information into commands in a very short time? How can these commands be quickly given to teenagers to control their actions: move forward quickly, cross the road, and avoid danger. Therefore, the mechanism by which the brain nerves acquire, transmit, and convert the electrical signals of nerves based on capabilities of human visual intelligence{\bf\color{purple}\cite{george01}} and the consistency of human activities for AI research is especially important.
\subsection{Knowledge graph-driven reasoning and fast decision-making}
{The knowledge graph is a large-scale semantic network composed of entities, concepts, attributes and the semantic relationships between them. Knowledge graph is actually a structured semantic knowledge base, which is used to describe the physical world in a symbolic form. Concepts and their interrelationships. Its basic unit is the {\texttt{Entity, Relationship, Entity}} triple, and the entity and its related attribute value pairs. The entities are interconnected through relationships., Constitutes a network of knowledge structure. Knowledge graph is a kind of knowledge base representing {\texttt{target$\cdot$concept$\cdot$entity and its relationship}} in the form of graph. This article discusses knowledge graph and knowledge respectively ontology concept, {\texttt{Ontology}}\color{red}\footnote{{The ontology is expressed as: \bf{O=\{C, P, R, F, A, I\}}, where:~Concept C,~properties P(Properties),~relationship R, ~function F,~axiom A(Axiom),~instance I (Instance)}} {\color{black}can be regarded as the pattern layer and logical basis of the knowledge graph. The research results can be used as the basis for the research of knowledge graphs, and knowledge graphs are the instantiation of ontology. The application of attention mechanism has greatly enriched the representation ability of neural networks. ({\bf\color{blue}1})~static knowledge graph attention mechanism, the encoded knowledge graph is used to enhance the semantic information of the input sentence, so that the input can be better understood;~({\bf\color{blue}2})~dynamic knowledge graph attention mechanism, read the knowledge graph and the knowledge triples in it, and then use the information in it to generate a better answer. Static and dynamic knowledge graph attention mechanisms are used to absorb common sense knowledge, which is good for understanding The user requests a dialogue with the generation. Generate high-concept dynamic knowledge representing {\texttt{entity-relation-entity}}~(ERE) and {\texttt{entity-attribute-value}}~(EAV) knowledge by "watching" the video figure, using visual commonsense reasoning~(VCR)~from recognition to cognition to build a combination of visual language model and static ontology tree to explain the working space, configuration, function and usage of humans and robots. This method is flexible, more general, and more Good robustness, suitable for various applications of autonomous robots.}}
\begin{figure*}[!ht]
\centering
\includegraphics[width=12.8cm,height=7.8cm]{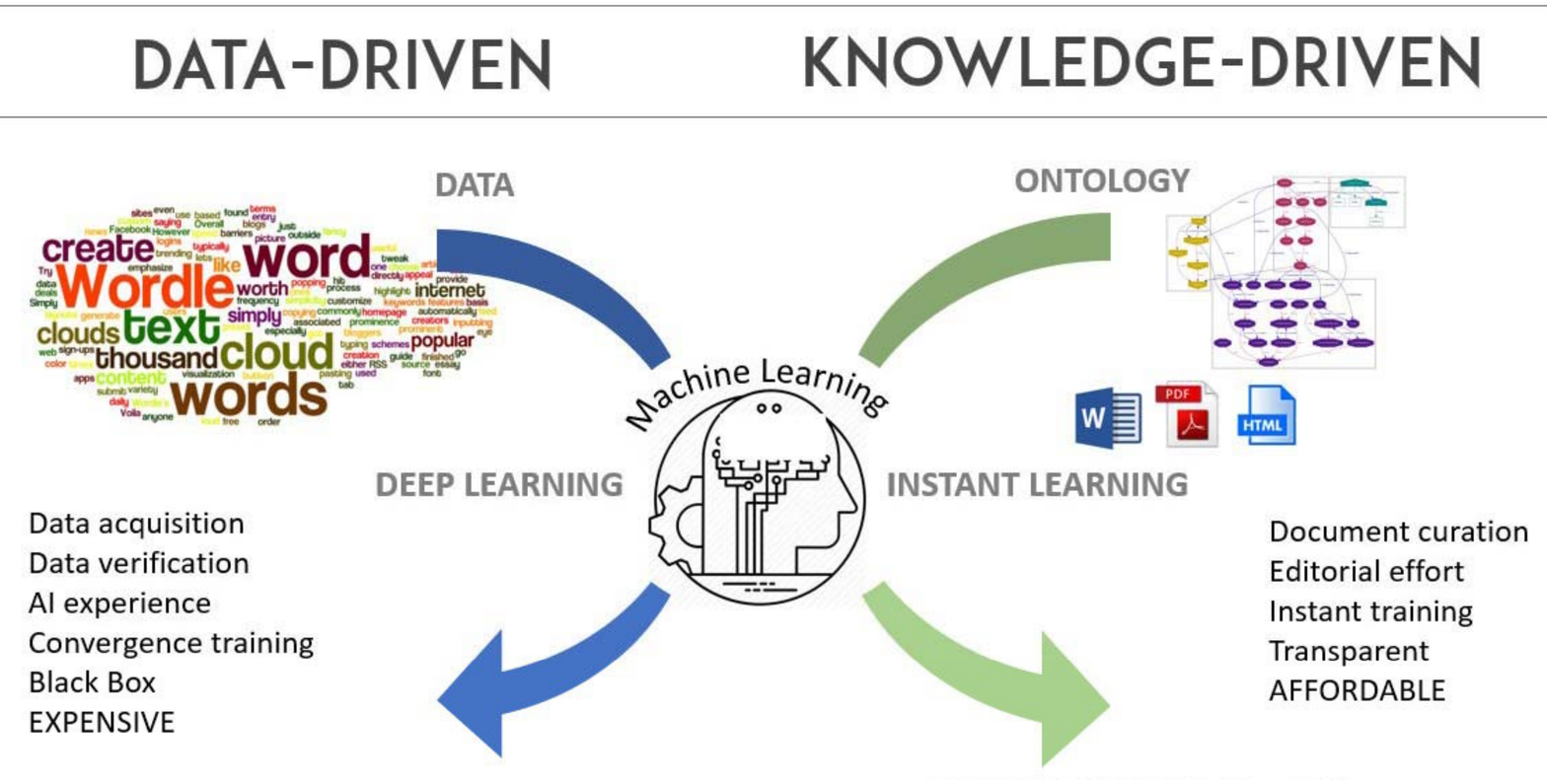}
\caption{{\bf{Data-driven and knowledge-driven machine learning.}} Includes deep learning and instant learning~(cited from Science, 2018)}
\label{fig06}
\end{figure*}
\section{Conclusion}
Artificial intelligence, computational intelligence, perceptual intelligence, cognitive intelligence, and decision-making intelligence reflect the different development stages of machine intelligence. Today's deep learning with data-driven models is approaching the limit of perceptual intelligence for feature learning and layered feature extraction with unsupervised or semi-supervised learning. The most important abilities of artificial intelligence are based on knowledge rather than data. The development of {\texttt{symbolic artificial intelligence}}{\bf\color{purple}\cite{segler01}} requires knowledge, especially {\texttt{symbolic knowledge}}. In the post-deep learning stage, the core of artificial intelligence should be {\texttt{knowledge representation}} and {\texttt{deterministic reasoning}}. Developing cognitive intelligence should enable machines to think like humans with the cognitive abilities to understand, explain, plan, reason, and make decisions, and to reach the advanced realm of human brain thinking. The realization of cognitive intelligence requires knowledge-driven or dual drive of data and knowledge (Fig.~{\bf\color{purple}\ref{fig06}}). To achieve the transformation of AI from perceptual intelligence to cognitive intelligence with human-level control, breakthrough in basic theoretical research on the frontier of AI is needed. From {\texttt{perceptual intelligence}} to {\texttt{cognitive intelligence}} is a disruptive leap in the development of artificial intelligence. Therefore, the analysis of brain cognitive functions, brain-inspired intelligent computational model construction, brain-inspired machine learning and meta-learning or learning to learn, interpretable advanced machine learning theories and methods, knowledge-driven commonsense intelligence, implementation of direct online communication between the machines and the brain, computational modeling and algorithms for complex deep convolutional neural networks research and innovation. We have been working on cutting-edge theoretical research on brain-inspired artificial intelligence computational modeling and new algorithms~{\emph{to expose the working mechanism of human brain, investigate brain-computer interface, explore new algorithms for deep meta learning and few-shot learning, build a knowledge-driven commonsense library, develop new technologies in AI, and create responsible AI systems}}. These challenging studies will be selected as the main research directions of AI scientists in the future.
%
\begin{figure*}[!htb]
\centering
\renewcommand \thefigure {S\arabic{figure}}
\centering
\includegraphics[width=17.8cm,height=7.2cm]{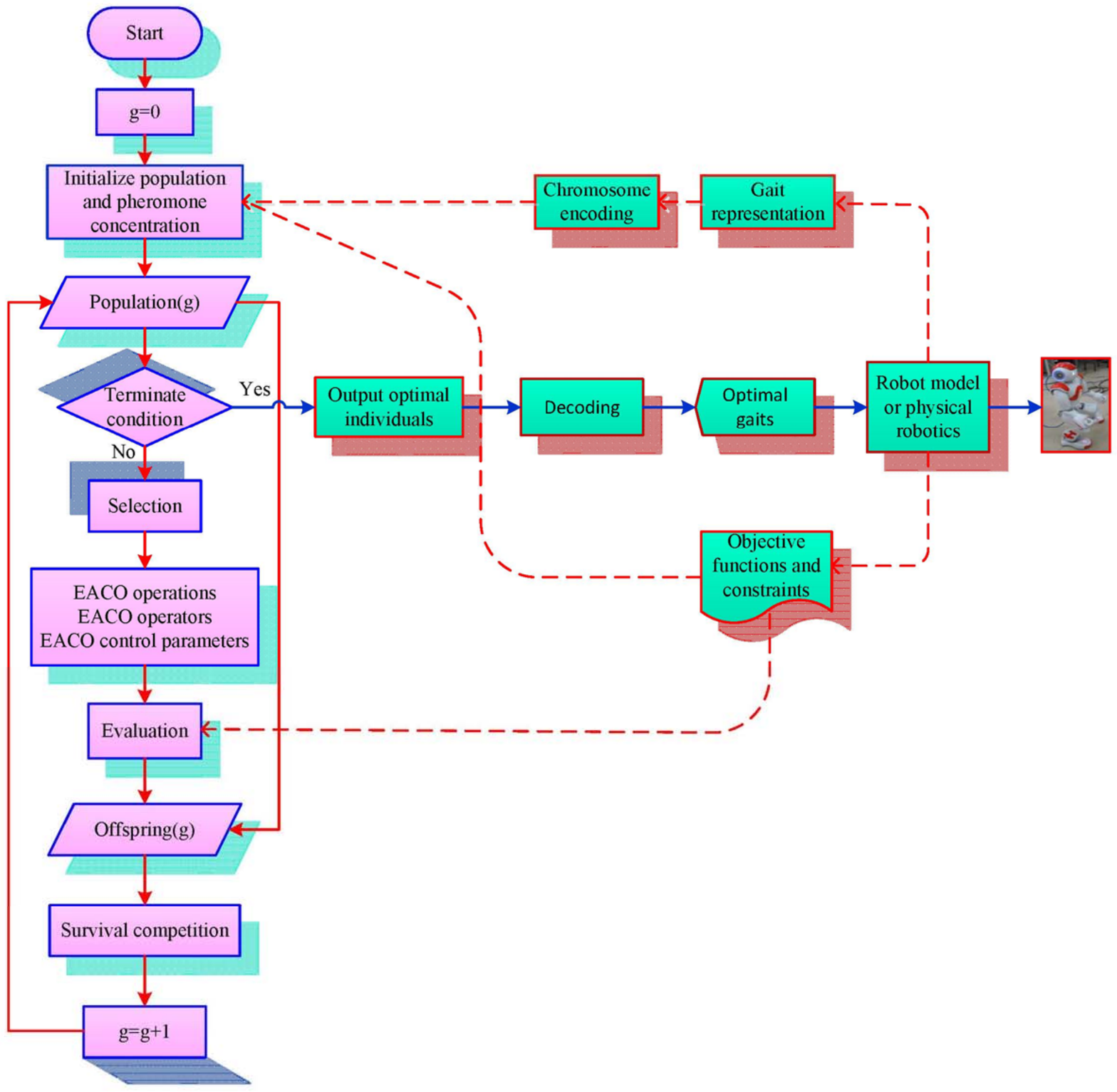}
\caption{Block diagram for obtaining optimal walking parameters for gait optimization of humanoid robotics based on EACO.}
\label{efig:01}
\end{figure*}
\bibliographystyle{ACM-Reference-Format}
\bibliography{yjaai.bbl}
\section*{Acknowledgments}
We thank Feng Wu of Changzhou Institute of Technology for assistance with coding and AILab at Stanford University for experimental and technical support. {\color{blue}{\bf{Funding}}}: This work is supported in part
by the Natural Science Foundation of China Grants no. 69985003. {\color{blue}{\bf{Author contributions}}}: J. Y. conceptualized the problem and the technical framework, developed and tested algorithm, designed and performed all experiments, collected and analyzed data, interpreted the results, created pictures and videos, and wrote and edited the paper. J. Y. and Y. P. managed the project. All authors have read and agreed to the published version of the manuscript. {\color{blue}\bf{Competing interests}}: All authors declare that they have no competing financial interests.
\noindent{\color{blue}\bf{Methods and all data}} are present in the
main text and supplementary materials. Correspondence should be addressed to yja431008@gmail.com.
\end{document}